# One-class Steel Detector Using Patch GAN Discriminator for Visualising Anomalous Feature Map

Takato Yasuno[1], Junichiro Fujii[1], Sakura Fukami[1]

[1] Research Institute for Infrastructure Paradigm Shift, Yachiyo Engineering Co., Ltd.,
Tokyo Taito Asakusabashi 5-20-8, Japan
{tk-yasuno,jn-fujii,sk-fukami}@yachiyo-eng.co.jp

**Abstract.** For steel product manufacturing in indoor factories, steel defect detection is important for quality control. For example, a steel sheet is extremely delicate, and must be accurately inspected. However, to maintain the painted steel parts of the infrastructure around a severe outdoor environment, corrosion detection is critical for predictive maintenance. In this paper, we propose a general-purpose application for steel anomaly detection that consists of the following four components. The first, a learner, is a unit image classification network to determine whether the region of interest or background has been recognised, after dividing the original large sized image into 256 square unit images. The second, an extractor, is a discriminator feature encoder based on a pre-trained steel generator with a patch generative adversarial network discriminator (GAN). The third, an anomaly detector, is a one-class support vector machine (SVM) to predict the anomaly score using the discriminator feature. The fourth, an indicator, is an anomalous probability map used to visually explain the anomalous features. Furthermore, we demonstrated our method through the inspection of steel sheet defects with 13,774 unit images using high-speed cameras, and painted steel corrosion with 19,766 unit images based on an eye inspection of the photographs. Finally, we visualise anomalous feature maps of steel using a strip and painted steel inspection dataset.

## 1 Introduction

### 1.1 Background and related works

**Manufacturing steel products and visual defect detection**. There have been many related studies on inspecting the manufacturing of steel products in indoor factory settings. The type of steel products can be categorised as billet, hot strip, cold strip, stainless steel, and rod/bar. A variety of defects can be listed according to the type of steel product [1]. For example, hot strip type steel sheets are extremely delicate and their surface should be inspected for quality control to determine whether there are any defects, such as rolled-in scales, patches, crazing, pits, inclusions, or scratches.

This study focuses on steel sheet defects. Studies on steel surface detection methods during the past 30 years can be classified into four groups: statistical, spectral filtering, model-based, and machine learning [1–3]. The use of a support vector machine, one of the mainstream supervised classification methods, has increased sharply since 2010. However, we have been unable to find any articles using a one-class support vector machine for steel surface detection. This study proposes a one-class detection method to determine whether the steel surface includes anomalous features or not, embedded by an input feature of a patch GAN discriminator. Youkachen et al. [4] applied a convolutional auto-encoder to reconstruct hot-rolled strip images, and the reconstructed images were then used to highlight the shape feature through simple post-processing algorithms. However, it is not yet widely understood whether the auto-encoder method can be beneficial to the problem of steel defect detection. This paper proposes an auto-encoder generator with a patch GAN discriminator whether real normal image or not normal (anomalous) one, in order to be more general application for steel anomaly detection.

**Vision-based inspection and health monitoring of steel structure**. There have been a number of studies on the inspection and health monitoring of the steel parts of infrastructure in severe outdoor environments. For example, several steel parts such as beams, slabs, and rivets are monitored for determining the health of steel bridges. For maintaining the painted steel surfaces of different types of infrastructure, steel damage detection is critical for predictive maintenance, such as rivet corrosion and rust and fatigue cracks in painted steel components.

Vision-based infrastructure inspection techniques have been researched mainly through supervised deep learning algorithms such as image classification, object detection, and semantic segmentation [5–7]. However, from a supervised learning standpoint, the classes of damage is a rare event and datasets including such events are always imbalanced, and hence, the number of rare class images is often relatively small. If deteriorations of infrastructure are progressing, then their events are not frequently occurred to be difficult to collect their anomalous images. Owing to this scarcity of damage data, it is difficult to improve the accuracy of supervised learning in an infrastructure inspection, which is one of the hurdles to overcome our underlying problems using damage detection for data mining based only on supervised learning approaches. Instead, this paper proposes a more general application for steel anomaly detection that consists of a supervised classification algorithm and an unsupervised anomaly detection algorithm embedded with an input feature through an adversarial auto-encoder network using a patch GAN discriminator.

### 1.2 Steel Anomaly Detector for Products and Maintenance

As illustrated in **Fig. 1**, in this paper, we propose a more general application for steel anomaly detection that consists of the following four components. The first component, is a unit image classification network learner for recognising the region of interest (ROI) or background, after dividing the original large size image into 256 pixels-square unit images. The second component, an extractor, is a discriminator feature encoder based on a pre-trained steel generator with a patch generative adversarial network discriminator. The third, an anomaly detector, is a one-class support vector machine used to predict the anomaly score using the input activated by the discriminator feature. The fourth indicator is an anomalous probability map used to explain the anomalous feature visually.

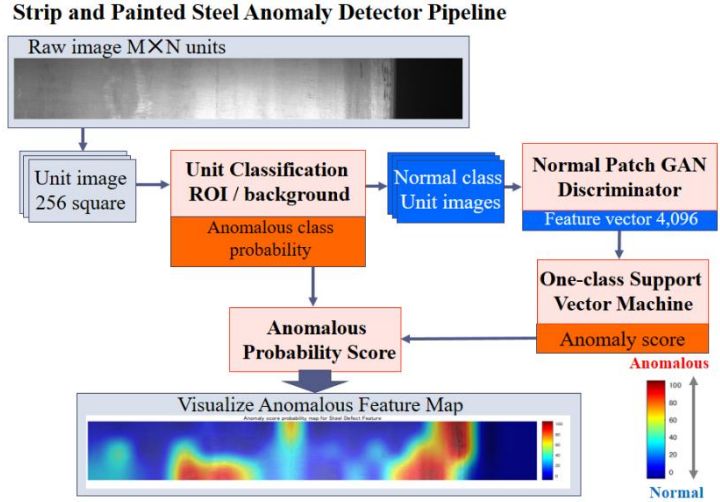

**Fig. 1** Overview of our pipeline for steel anomaly detection, showing the raw input divided into smaller sized 256 pixels-square unit images. The resulting unit images are further classified to determine whether they show the background or steel region of interest, including the normal class and anomalous class. Herein, the normal class unit images are generated into a fake steel output to fool the patch GAN discriminator. Furthermore, the feature vector from the final layer of the patch GAN discriminator is machine learned using a one-class SVM to compute a high anomaly score from the normal features. Finally, we can visualise the anomalous feature map multiplied by two predicted outputs of the anomalous class probability from the pre-trained classification and the anomaly score from the pre-trained one-class SVM model.

## 2   Steel Anomaly Detection Application

Our proposed steel anomaly detection method contains the following components: 1) a unit image classification network, 2) a steel generator with a patch GAN discriminator, 3) a one-class SVM to predict the anomaly score using the discriminator feature, and 4) an anomalous feature map to visualise the anomalous features.

### 2.1   Unit image classification among ROI and background

The first learner is a unit image classification network to determine whether we recognise the ROI or background, after dividing the original raw image $I$ with a height of 1,500 pixels and width of 2,000 pixels into 256 pixels-square unit images $U_i (\text{i} = 1, \ldots, \text{M} \cdot \text{N})$ with M rows and N columns. We can train classification models using several candidates of deep neural networks with the advantage of a higher test accuracy and normal and anomalous accuracy classes for the target ROI and background of steel sheets and painted steel. We propose and compare usable deep neural networks such as EfficientNet-b0 [8], Xception-v2 [9], Inception-v3 [10], ShuffleNet [11], and Res-Net101 [12]. We can evaluate their test accuracy with both recall and precision to compute their confusion matrix. For steel anomaly detection used to maintain the quality of manufacturing products, and structural health monitoring of the status of steel bridge parts, the recall is more important than the precision. For this reason, we should avoid an improper inspection despite it containing anomalous features for quality control and predictive maintenance.

### 2.2   Feature extractor using a patch GAN discriminator

The second extractor is a discriminator feature encoder based on a pre-trained steel generator with a patch GAN discriminator. There are several anomaly detectors using a GAN, including AnoGAN [13], GANomaly [14], and skip-GANomaly [15]. We have found that these state-of-the-art medical image [16] and baggage security [17] approaches frequently generate artifacts with different features from the target region of interest when applied to material surface in concrete infrastructure: bridge, dam. For this reason, we customised a generative deep network for steel anomaly detection.

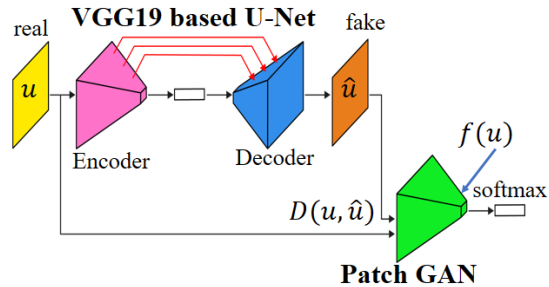

**Fig. 2** VGG19 based U-Net generator with a patch GAN discriminator

As illustrated in **Fig. 2**, we formulate a customised network that consists of two depth of encoder-decoder layer with the standard backbone of VGG19 [18], where skip-connected from encoder to decoder at the same scale to concatenate their channels inspired with the U-Net [19]. It also consists of a patch GAN discriminator to determine whether a real normal image or a fake (not real) image (not normal and anomalous) output created from the generator. The generator is a U-Net style generator with 46 layers with the encoder of a VGG19, an over-complete decoder with four-times the number of channels, and bridge layers with three blocks of convolution and parametric ReLU layers. There is also a dropout layer before the decoder block respectively. By contrast, the discriminator is a patch GAN encoder with four down sampling and has 17 layers consisting of a convolution, ELU, and fully connected layer at the final activation. While training these deep networks, the generator attempts to fool the discriminator and simultaneously the discriminator learn to determine whether the generated output is real normal image or not. For the third anomaly detector, we can extract a discriminator feature at the final convolution layer $f(U_i)(i = 1, ..., M \cdot N)$, with 4,096 learnable values, transformed into the feature vector for the input of the one-class support vector machine.

### 2.3  Anomaly scorer using one-class support vector machine

The third anomaly detector is a one-class SVM used to predict the anomaly score using the activated discriminator feature. Under the one-class SVM, the less negative the anomaly score $v(f(U_i))(i = 1, ..., M \cdot N)$ is, the higher the anomalous features. Without a loss of generality, we can reverse the sign of the normalised anomaly score such that the more positive the score is, the higher anomalous features are as follows:

$$score(U_i) = (-1) \cdot (v(f(U_i)) - \min v)/(\max v - \min v) \qquad (1)$$

Here, $\max v$ represents the maximum anomaly score using the one-class SVM. By contrast, $\min v$ stands for the minimum anomaly score using it. Note that the negative sign is for the reverse of the anomalous measure, and the more positive $score(U_i)$ is, the higher the anomalous features for a relevant unit image.

### 2.4  Anomalous feature map visualising steel defects and corrosion

The fourth indicator is an anomalous probability map used to visualise the anomalous features. Let us indicate that $U_i(i = 1, ..., M \cdot N)$ is the i-th unit image with 256 pixels-square, among which the raw image is divided into unit images of M rows multiplied by N columns. The anomalous feature $AF(U_i)$ of each unit image is represented as the next function that multiplies the anomalous class probability and the anomaly score.

$$AF(U_i) = \sum_{a \in A} prob_a(U_i) \cdot score(U_i), (i = 1, ..., M \cdot N) \qquad (2)$$

Here, $a \in A$ is one of the anomalous class sets among all classes that includes the ROI and background over the unit classification network. The anomalous class probability $prob_a(U_i), a \in A$ takes a range of between 0 and 1. Thus, the computation of Equation (2) enables to visualise the anomalous feature map, thereby transforming the original sized heat map composed of M rows multiplied by N columns with the unit anomalous feature, and to overlay them on the relevant raw image.

## 3  Applied Results

We demonstrate our method on inspecting strip steel sheet defects using high-speed camera images and painted steel corrosion through photographs inspected by the human eye. To achieve a more stable pipeline, we compare the classification networks and analyse their accuracy.

### 3.1 Dataset of steel image detection for products and monitoring

As shown in **Table 1**, we prepared two datasets of camera images of strip steel sheets [20] and eye-inspected photographs of painted steel slab parts. The former dataset is for a 2019 competition conducted by Severstal Co., where the number of training data is 12,000 and the number of test data is 5,000. In this study, we randomly sampled 2,000 training data and 30 test data with a raw image size with a height of 256 and a width of 1,600. The original width of the raw images was scaled up to 1,792, which is equal to 256 multiplied by 7 without any remainder. Therefore, 7 unit images per original raw image were created to form 14,000 unit images, and after cleansing, 13,774 were usable. The latter dataset was provided to us solely for collaborative research by the sub-working group at the Public Works Research Institute (PWRI) in Japan. Although the angle and distance toward the bridge slab were different, 256 pixels-square unit images were created after cleansing, and 19,766 were usable, excluding those having a pixel size of less than 256 at the edge of raw image.

**Table 1.** Dataset of camera images of steel sheets and eye-inspection photographs of bridges.

| Dataset example (situation) | Target of steel damage feature definition of classification | Raw image size number of unit clip |
|---|---|---|
| Manufacturing steel products [20] (indoor) | Strip steel defect 6-classes: 1) normal, and anomalous 4-classes: 2) rolled-in scale, 3) scratch, 4) patch, 5) inclusion, 6) background. | H256×W1,600 $M \cdot N = 13,774$ |
| Bridge eye inspection: RC slab part (outdoor) | Painted steel damage 7-classes: 1) normal painted steel, 2) corrosion, 3) concrete (CO.), 4) mixed CO. and background, 5) mixed CO. and painted steel, 6) dark, 7) background. | H1,536×W2,048 $M \cdot N = 19,766$ |

**Table 2.** Comparison of classification network for painted steel corrosion
(test accuracy of all seven classes and precision of four key classes)

| classification model | 7class test accuracy | key class precision | | | | |
|---|---|---|---|---|---|---|
| | | 4class precision | Paint steel | Corrosion | mixed ST-CO | Concrete |
| EfficientNet-b0 | 0.865 | 0.879 | 0.838 | **0.874** | 0.853 | 0.949 |
| Xception-v2 | 0.864 | 0.878 | **0.870** | 0.836 | **0.862** | 0.942 |
| Inception-v3 | 0.864 | 0.876 | 0.863 | 0.831 | 0.858 | 0.952 |
| ShuffleNet | 0.868 | 0.877 | 0.819 | 0.863 | **0.867** | **0.960** |
| ResNet101 | **0.889** | **0.892** | 0.865 | **0.890** | 0.860 | **0.954** |

### 3.2 Trained unit image classification

As indicated in **Table 2**, we trained five candidate classification networks for painted steel corrosion detection to compare their test accuracy and class precision. Here, we set the ratio of training versus testing at 80:20. The test accuracy of ResNet101 with the highest score of 0.889 outperformed the other four networks. Considering the key class precision for painted steel, corrosion, mixed steel and concrete, and concrete, the four-class precision achieves the highest score of 0.892 for ResNet101.

**Table 3.** Classification networks for painted steel (continue: recall of 4-classes and runtime)

| classification model | key class recall | | | | | runtime |
|---|---|---|---|---|---|---|
| | 4class recall | Paint steel | Corrosion | mixed ST-CO | Concrete | |
| EfficientNet-b0 | 0.873 | 0.883 | 0.848 | 0.808 | **0.952** | 269m |
| Xception-v2 | 0.870 | 0.847 | **0.872** | 0.806 | **0.956** | 200m |
| Inception-v3 | 0.871 | 0.850 | **0.887** | 0.794 | **0.952** | 286m |
| ShuffleNet | 0.878 | **0.895** | 0.846 | **0.822** | 0.947 | 49m |
| ResNet101 | **0.890** | **0.894** | 0.859 | **0.857** | 0.950 | 309m |

As shown in **Table 3**, we confirmed the training results of the key class recall and runtime. The recall of the four key classes also achieved the highest score of 0.890 for ResNet101. Thus, we selected ResNet101 as the unit image classification network for painted steel corrosion detection. **Table 4** describes the confusion matrix of a bridge eye inspection under the most accurate ResNet101 with a test accuracy of 88.9%.

**Table 4.** Confusion matrix of bridge eye inspection for ResNet101 with a test accuracy 88.9%.

RC slab 7-Classification Confusion Matrix

| | c1CoROI | c2mixCoBack | c3Background | c4PaintSteel | c5mixCoPaint | c6DarkCoUnused | c7StCorrosion | | |
|---|---|---|---|---|---|---|---|---|---|
| c1CoROI | 1069 | 6 | 1 | 3 | 33 | | 9 | 95.4% | 4.6% |
| c2mixCoBack | 1 | 91 | 10 | 4 | 1 | | 3 | 82.7% | 17.3% |
| c3Background | 1 | 9 | 235 | 3 | 2 | 2 | 9 | 90.0% | 10.0% |
| c4PaintSteel | 11 | 1 | 8 | 658 | 37 | 11 | 35 | 86.5% | 13.5% |
| c5mixCoPaint | 28 | 1 | 3 | 24 | 704 | 7 | 52 | 86.0% | 14.0% |
| c6DarkCoUnused | 5 | 2 | 4 | 13 | 11 | 81 | 3 | 68.1% | 31.9% |
| c7StCorrosion | 10 | | 8 | 31 | 33 | 2 | 678 | 89.0% | 11.0% |
| | 95.0% | 82.7% | 87.4% | 89.4% | 85.7% | 78.6% | 85.9% | | |
| | 5.0% | 17.3% | 12.6% | 10.6% | 14.3% | 21.4% | 14.1% | | |

**Table 5.** Revised training of classification network for manufacturing of steel products (test accuracy, class precision, and the lowest precision for 6 classes)

| classification model | 6class test accuracy | key class precision | | | | | | lowest precision |
|---|---|---|---|---|---|---|---|---|
| | | normal | 4class precision | rolled-in scale | inclusion | scratch | patch | |
| ResNet101 1,774 | 0.775 | **0.748** | 0.720 | 0.857 | **0.689** | 0.714 | 0.619 | 0.619 |
| ResNet101 5,774 | **0.819** | **0.794** | 0.784 | **0.937** | **0.692** | **0.722** | 0.784 | **0.692** |
| ResNet101 9,774 | 0.807 | 0.741 | 0.783 | 0.894 | 0.641 | **0.772** | 0.823 | 0.641 |
| ResNet101 13,774 | 0.804 | 0.723 | 0.777 | 0.873 | 0.663 | 0.706 | **0.867** | **0.663** |

As indicated in **Table 5**, we used a four-step trained classification network, ResNet101, to increase the number of training data for the detection of manufactured steel to 1,774, 5,774, 9,774, and 13,774. We set the ratio of training versus testing to 80:20. The test accuracy of ResNet101 achieved a stable score of 0.804. Considering the key class precision, which includes rolled-in scales, inclusions, scratches, and patches, the four anomalous class precision enabled a stable score of 0.777 to the lowest precision of 0.663 when 13,774 unit images were learned by ResNet101.

**Table 6.** Revised training classification network for manufacturing of steel products (class recall and the lowest recall)

| classification model | key class recall | | | | | | lowest recall |
|---|---|---|---|---|---|---|---|
| | normal | 4class precision | rolled-in scale | inclusion | scratch | patch | |
| ResNet101 1,774 | **0.896** | 0.702 | 0.583 | 0.718 | 0.694 | **0.812** | 0.583 |
| ResNet101 5,774 | 0.891 | **0.791** | 0.655 | **0.901** | 0.812 | **0.795** | 0.655 |
| ResNet101 9,774 | **0.897** | 0.780 | **0.675** | 0.832 | 0.827 | 0.785 | **0.675** |
| **ResNet101 13,774** | 0.844 | **0.789** | **0.708** | 0.810 | **0.874** | 0.765 | **0.708** |

As shown in **Table 6**, we confirmed the training results of the key class recall and runtime. The recall of the four key classes enable a stable score from 0.789 to the lowest recall of 0.708 under ResNet101. Therefore, the more the number of training data increased, the higher the accuracy of the bottom level among the key classes, that is, the lowest precision and recall. **Table 7** describes the confusion matrix of manufacturing steel products using the updated ResNet101 with a test accuracy of 80.4%.

**Table 7.** Confusion matrix of manufacturing steel products under the revised ResNet101 with a test accuracy of 80.4%, in case of the number of dataset 13,774.

Steel Defect Classification Confusion Matrix

| | c1normal | c2urokoScale | c3inclusion | c4scratch | c5patch | c6background | | |
|---|---|---|---|---|---|---|---|---|
| c1normal | 222 | 70 | 7 | 1 | | 7 | 72.3% | 27.7% |
| c2urokoScale | 16 | 345 | 17 | 4 | 13 | | 87.3% | 12.7% |
| c3inclusion | 16 | 42 | 128 | 5 | | 2 | 66.3% | 33.7% |
| c4scratch | 6 | 25 | 6 | 125 | 11 | 4 | 70.6% | 29.4% |
| c5patch | | 5 | | 7 | 78 | | 86.7% | 13.3% |
| c6background | 3 | | | 1 | | 210 | 98.1% | 1.9% |
| | 84.4% | 70.8% | 81.0% | 87.4% | 76.5% | 94.2% | | |
| | 15.6% | 29.2% | 19.0% | 12.6% | 23.5% | 5.8% | | |

### 3.3 Trained one-class SVM using patch GAN discriminator feature

We trained a steel surface generator with a patch GAN discriminator using normal steel surface images numbering 2,070. Here, the input size is 100 pixels-square with three channels, and we set the mini-batch size to 4 and implemented 8 epoch iterations. Furthermore, we trained a one-class SVM embedded by the input of our trained patch GAN discriminator feature with 4,096 elements. Herein, the training data was made up of 2,070 normal steel surface unit images, and the test data included 1,000 normal steel images and 900 anomalous steel surface images of four classes containing 400 rolled-in scales, 200 inclusions, 200 scratches, and 100 patches.

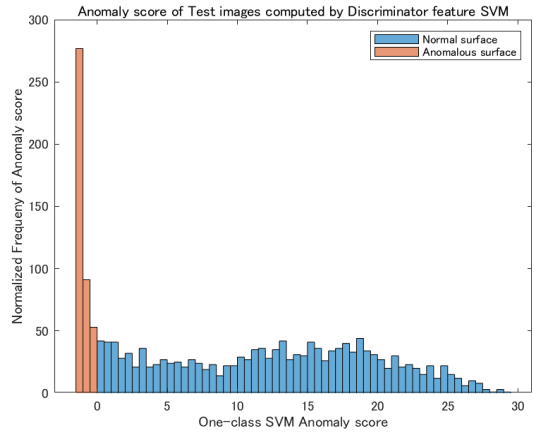

**Fig. 3** Histogram of anomaly score for manufacturing steel products

Furthermore, **Fig. 3** shows a histogram of the anomaly score for manufacturing steel products using our trained one-class SVM model embedded into the input of our trained patch GAN discriminator feature. On the left-side of the brown colour bar, the negative value of the anomaly score stands for the anomalous steel surface prediction using the one-class SVM. On the right-side of the blue colour bar, the positive value of anomaly score represents the normal steel surface prediction. The left-side of **Fig. 4** depicts the most anomalous steel surface of 100 unit images for manufactured steel products. Remarkably, there are anomalous classes of steel defect images that contain rolled-in scales, inclusions, scratches, and patches. By contrast, the right-side of **Fig. 4** shows the most normal 100 unit images without any steel defects.

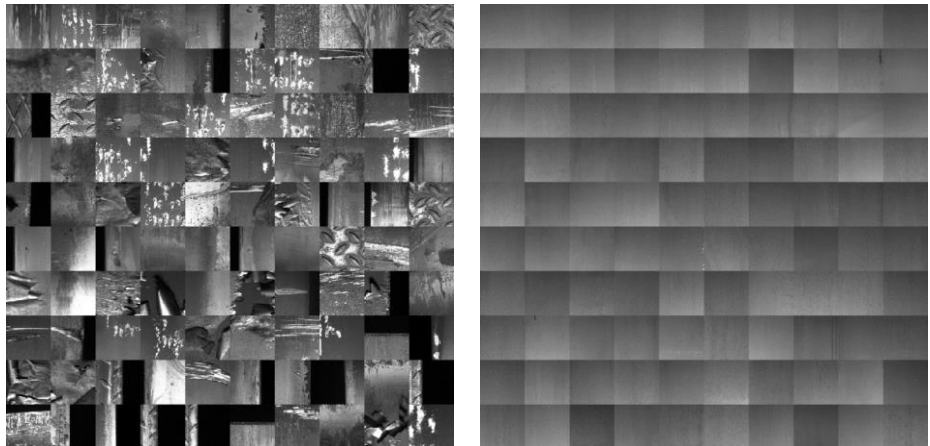

**Fig. 4** The most anomalous 100 images (left) using the highest score for manufactured steel products, and the most normal 100 images (right) using the lowest score.

In addition, we trained a painted steel generator with a patch GAN discriminator using 5,504 normal painted steel images of a bridge for eye inspection. Here, the input size is 100 pixels-squares with three channels, and we set the mini-batch size to 4 and implemented 8 epoch iterations. Furthermore, we trained a one-class SVM embedded by the input of our trained patch GAN discriminator feature with 4,096 elements. Herein, the training data is 2,000 normal painted steel unit images, and the test data includes 895 normal painted steel images and 905 anomalous painted steel surface images with a corrosion feature. Furthermore **Fig. 5** shows a histogram of the anomaly score for a bridge eye inspection using our trained one-class SVM model embedded into the input of our trained patch GAN discriminator feature. On the left-side of the brown colour bar, the negative anomaly score stands for the anomalous painted steel surface prediction under the one-class SVM. On the right-side of the blue colour bar, the positive value of the anomaly score represents the normal painted steel surface.

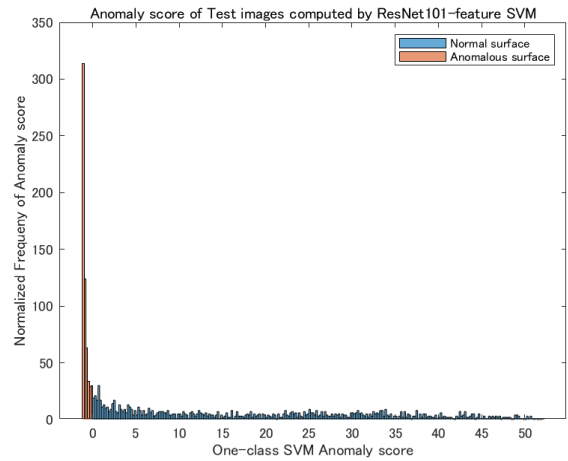

**Fig. 5** Histogram of anomaly score for bridge eye inspection

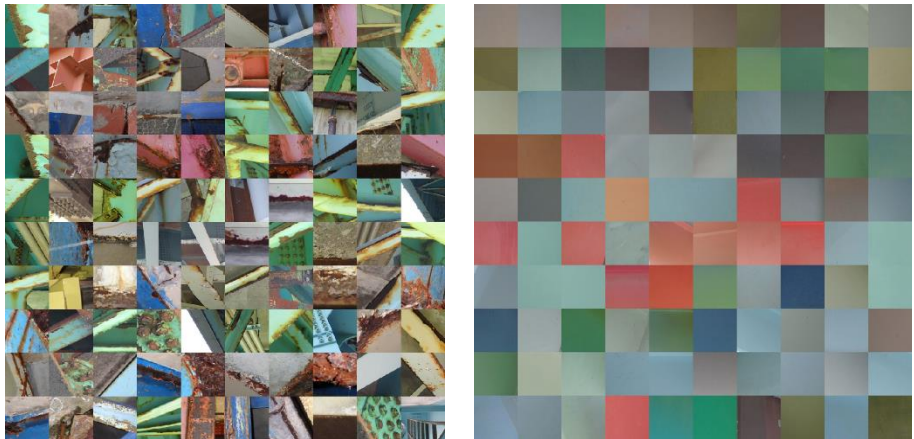

**Fig. 6** The 100 most anomalous images (left) using anomaly score for bridge eye inspection, and the 100 most normal images (right) using the anomaly score.

The left-side of **Fig. 6** depicts the 100 most anomalous painted steel surface unit images of a bridge slab for a steel part inspection. Severely, there are anomalous painted steel surface corrosion images that contain deteriorated beams and rivets. By contrast, the right-side of **Fig. 6** shows the 100 most normal unit images without any steel corrosion.

### 3.4 Visualisation of anomalous feature map

As shown in **Fig. 7**, we can visualise examples of anomalous feature maps for manufacturing steel products. These are top and bottom ten pairs of raw steel sheet images and anomalous feature maps whose montage consists of five rows and two columns. The black region shows the background without any steel sheet region. On the anomalous feature map, the red colour represents anomalous features at those pixels where the anomalous probability score is high at around 100. By contrast, the blue pixels indicate that the anomaly probability score is low at close to zero. Focusing on the region of steel defects such as rolled-in scales, inclusions, scratches, and patches, these anomalous feature maps are able to enhance the red pixels when overlaid onto the raw input of the steel sheet surface image.

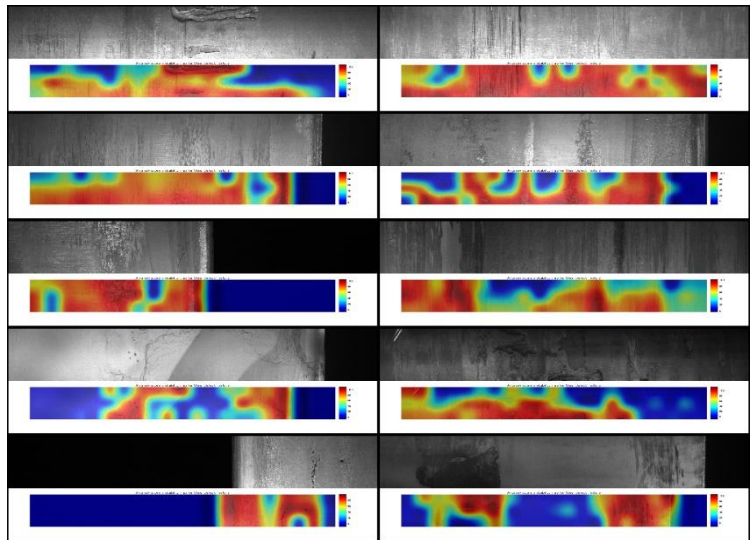

**Fig. 7** Anomalous feature map for manufacturing steel products (five rows and two columns)

As shown in **Fig. 8**, we can visualise cases of anomalous feature maps for bridge slab photographs using a human eye inspection. These are ten pairs of left and right raw bridge slabs that consist of beams, concrete, painted steel parts, and rivets and anomalous feature maps whose montage consists of five rows and two columns. On the right side of the anomalous feature map, the red represents anomalous feature at those pixels where the anomalous probability score is high at approximately 100. By contrast, over the blue-to-red (jet) colour space, the pixels in blue indicate that the anomaly probability score is low close to zero. In the region of steel corrosion, these anomalous feature maps are able to emphasise red colour pixels overlaid onto the raw input of a bridge slab image.

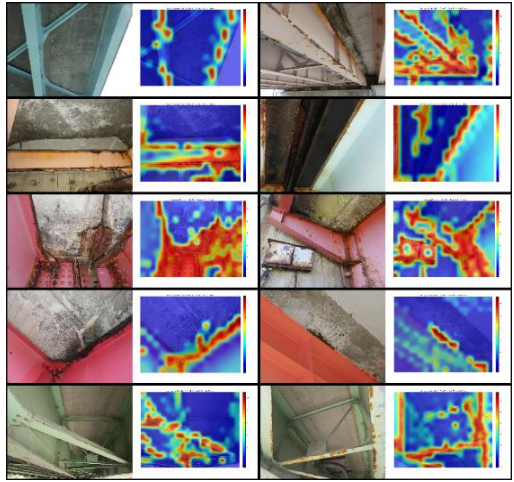

**Fig. 8** Anomalous feature map for bridge eye inspection (cases of five rows and two columns)

## 4     Concluding Remarks

### 4.1     Steel anomaly detection pipeline for products and monitoring

This study proposed a more general application for steel anomaly detection that consists of the following four components: 1) unit image classification network to determine whether we recognise the ROI or background, 2) a steel generator with a patch GAN discriminator, 3) a one-class SVM to predict the anomaly score using the discriminator feature, and 4) an anomalous probability map to visualise the anomalous features. Furthermore, we demonstrated our method for inspecting steel sheet defects with 13,774 unit images using high-speed cameras, and painted steel corrosion with 19,766 unit images using eye inspection photographs. Therefore, we visualised steel anomalous feature maps using the strip and painted steel inspection datasets. Focusing on the region of steel defects such as rolled-in scales, inclusions, scratches, and patches, these anomalous feature maps can enhance the red colour pixels overlaid onto the raw input of strip steel surface images. In addition, regarding a region of painted steel corrosion over a beam and rivet, these anomalous feature maps can emphasise red colour pixels overlaid onto the raw input of bridge slab images.

### 4.2     Future studies for robust location and various interest

In the future, we will continue to learn another dataset of various materials such as concrete and asphalt. The location of the infrastructure can be different, e.g., reinforcement of concrete parts on bridge slabs, and asphalt or concrete on road pavement. Although we focused on manufactured products and bridge maintenance, fur-ther cases can be applied, such as after a natural disaster including an earthquake or flood, which frequently occur in Japan with devastating impact. The accuracy of our pipeline mainly depends on the unit image classification to divide the region of interest or background, and a one-class SVM using a normal patch GAN discriminator. Using another dataset of the different location and various region of interest, the parameters of classification are able to adapt to yet unseen normal features, anomalous features, and the background. Although we learned a U-Net stylised generator using the backbone of VGG19, another backbone network can be revised for a more efficient and stable detection.

**Acknowledgments** The authors would like to thank Takuji Fukumoto and Shinichi Kuramoto (MathWorks) for providing MATLAB resources for generative learning.